\pdfoutput=1

\documentclass[11pt]{article}

\usepackage[final]{acl}

\usepackage{times}
\usepackage{latexsym}

\usepackage[T1]{fontenc}

\usepackage[utf8]{inputenc}

\usepackage{inconsolata}

\usepackage[acronym, nomain]{glossaries} 

\usepackage{array} 
\usepackage{booktabs}

\usepackage{microtype}
\usepackage[most]{tcolorbox}
\usepackage{pifont}
\usepackage{scrextend}
\usepackage{pifont}
\usepackage{tabularx}
\usepackage{booktabs}
\usepackage{graphicx}
\usepackage{subcaption}
\usepackage{xurl}
\usepackage{stmaryrd}
\usepackage{amsmath}
\usepackage{cleveref}
\usepackage{xcolor}
\usepackage{tikz-dependency}
\usepackage{flowchart}
\usepackage{pgf}
\usetikzlibrary{arrows.meta,shapes.symbols,shapes.geometric,shadows}
\usepackage{varwidth}
\usepackage{tabularx}
\usepackage{tablefootnote}
\usepackage{makecell}
\usepackage{multirow}
\usepackage{array}
\usepackage{arydshln}
\usepackage{lingmacros}
\usepackage{enumitem}
\usepackage{soul}

\glsdisablehyper 
\newacronym{sts}{STS}{Semantic textual similarity}
\newacronym{csts}{C-STS}{Conditional STS}
\newacronym{nlp}{NLP}{natural language processing}
\newacronym{llms}{LLMs}{large language models}
\newacronym{qa}{QA}{Question Answering}
\newacronym{tfs}{TFS}{typed-feature structure}

\title{Linguistically Conditioned Semantic Textual Similarity}

\author{Jingxuan Tu \and Keer Xu \and Liulu Yue \\
{\bf Bingyang Ye \and Kyeongmin Rim  \and James Pustejovsky} \\
        Department of Computer Science \\ Brandeis University \\ Waltham, Massachusetts, USA \\
        {\tt\{jxtu,keerxu,liuluyue,byye,krim,jamesp\}@brandeis.edu}
        }

\begin{document}

\maketitle

\begin{abstract}

\gls{sts} is a fundamental NLP task that measures the semantic similarity between a pair of sentences. 
In order to reduce the inherent ambiguity posed from the sentences, a recent work called \gls{csts} has been proposed to measure the sentences' similarity conditioned on a certain aspect.
Despite the popularity of \gls{csts}, we find that the current \gls{csts} dataset suffers from various issues that could impede proper evaluation on this task. 
In this paper, we reannotate the \gls{csts} validation set and observe an annotator discrepancy on 55\% of the instances resulting from the annotation errors in the original label, ill-defined conditions, and the lack of clarity in the task definition.
After a thorough dataset analysis, we improve the \gls{csts} task by leveraging the models' capability to understand the conditions under a QA task setting. 
With the generated answers, we present an automatic error identification pipeline that is able to identify annotation errors from the \gls{csts} data with over 80\% F1 score.
We also propose a new method that largely improves the performance over baselines on the \gls{csts} data by training the models with the answers.
Finally we discuss the conditionality annotation based on the \gls{tfs} of entity types.
We show in examples  that the \gls{tfs} is able to provide a linguistic foundation for constructing \gls{csts} data with new conditions.

\end{abstract}

\glsresetall
\section{Introduction}

\begin{figure}[h!]
  \centering
  \includegraphics[width=0.9\linewidth]{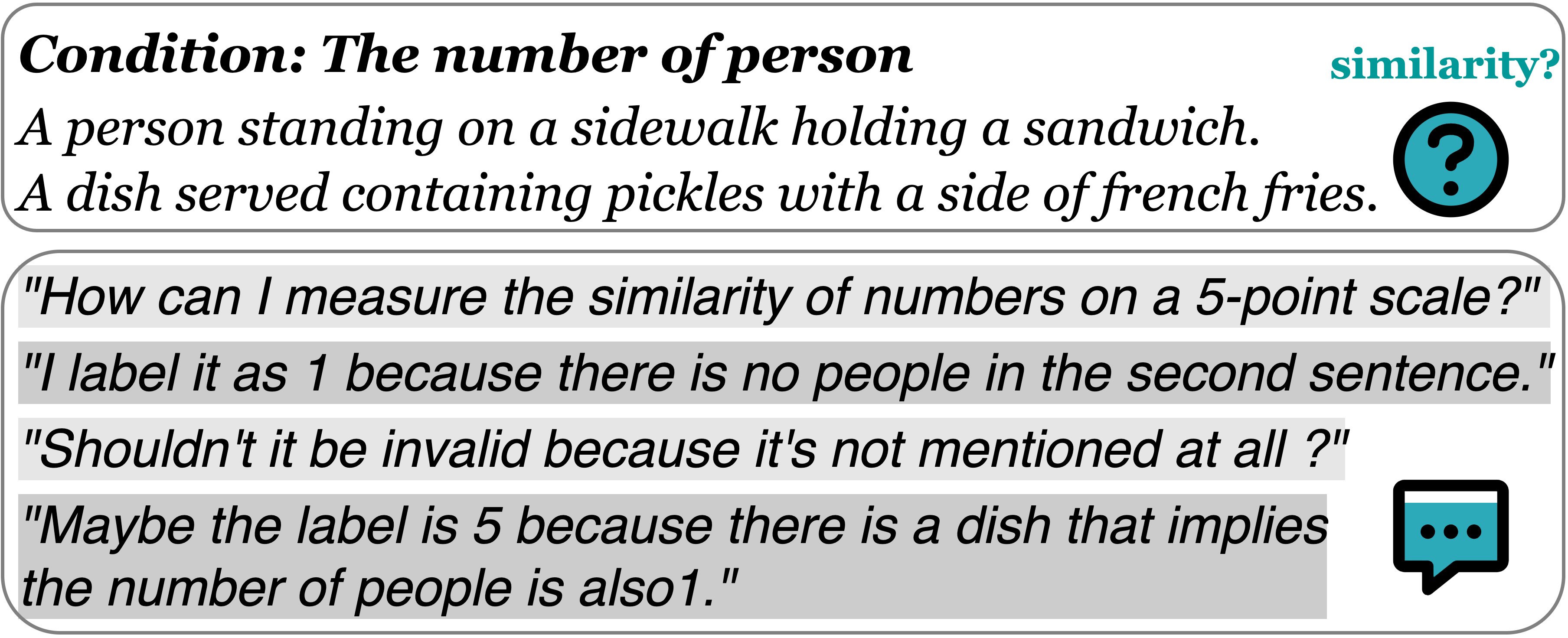}
  \caption{A problematic example from the C-STS dataset. The binarity of the condition cannot be mapped to a 5-point similarity scale. The label can be subjective depending on how much inference is made from the context. No guideline on the scenario when the information regarding the condition is missing.}
  \label{fig:example}
\end{figure}

\gls{sts} is an essential NLP task that measures the semantic similarity between two sentences \cite{sts12}.
It is also a popular benchmark for developing tasks such as text embedding learning \cite{senteval,sbert,sbert-aug} and language understanding \cite{glue}.
While the \gls{sts} datasets have been developed and improved over the past years \cite{sts13,sts14,sts15,sts16,sts17}, 
the task itself still suffers from sentence ambiguity and subjectivity to judgment \cite{csts}. 

A new task called \gls{csts} has been proposed to resolve those issues \cite{csts}. It is designed to disambiguate the similarity between two sentences by measuring the similarity on a given condition. 
An accompanying dataset was also proposed to test models on the \gls{csts} task. 
Despite the popularity of \gls{csts}, we observe certain limitations in the \gls{csts} dataset that could hinder the understanding and proper evaluation of models on this task.
As illustrated in Figure \ref{fig:example}, these limitations primarily revolve around annotation errors, ill-defined conditions, and a general lack of clarity in task definition. 

Taking into account the significance of these issues, we intend to improve the \gls{csts} dataset by addressing the existing problems that we observed.
We start by reannotating the \gls{csts} validation set. By identifying an apparent annotation error rate of 55\%\footnote{Calculated from the comparison between the original and reannotated labels.} in their validation set, we analyze the provenance of the errors and discrepancies between the original and relabeled datasets.

To further explore the utility of the condition and how it is understood by language models, we treat it as a \gls{qa} task and leverage \gls{llms} to generate the answer to the question that is constructed from the condition. 
We find that the LLM-generated answers can better capture the similarity between two sentences and fit closely to our reannotated labels by having a higher Spearman's Correlation. 
Based on this finding, we propose an approach to identify potential annotation errors from the \gls{csts} dataset utilizing the LLM-generated answers, achieving over 80\% F1 score on the validation set.
We also propose a new method to improve the \gls{csts} task by training the models with the answers. We show that both supervised and generative models can efficiently and effectively learn the condition information encoded in the answers, improving the performance over baselines by a large margin. 

Finally, we discuss a new annotation specification of the conditionality that aims to improve the formulation of the conditions with a more concrete semantic base.
We use the entity type identified from the sentence pair as the surface condition text that is described by its underlying \gls{tfs} \cite{carpenter1992logic,copestake2000definitions,penn2000algebraic}.
We exemplify that \gls{tfs}-based conditions can be successfully adopted to sentence pairs from the current \gls{csts} dataset.

We summarize the main contributions of this paper as threefold. We reannotate the \gls{csts} validation set and propose an error identification pipeline that can be applied to the whole dataset to identify potential annotation errors and ambiguities; we propose a \gls{qa}-facilitated method that largely improves the model performance on the \gls{csts} task; we discuss using \gls{tfs}
 as a new annotation specification to improve the conditionality in \gls{csts} dataset with a more concrete semantic base.
We make the source code and dataset publicly available.\footnote{\url{https://github.com/brandeis-llc/L-CSTS}}



\section{Related Work and Background}
\paragraph{Semantic similarity tasks}
The semantic similarity between texts has long been a key issue in understanding natural language.
\citet{sts12} proposed the first \gls{sts} task that measures the similarity between a sentence pair. Following this, \citet{sts13} proposed the second \gls{sts} task that covered more text genres in the dataset. 
There are also \gls{sts} tasks \cite{sts14,sts15,sts16,sts17} with focuses on measuring sentence similarity under a multilingual and cross-lingual setting. 
\citet{str} proposed a new text similarity task that measures the semantic relatedness between two sentences.

\paragraph{Conditional \gls{sts}} More related to our work, the \gls{csts} task \cite{csts} introduced conditions on top of the traditional \gls{sts} tasks, and it measured the sentence similarity regarding the given condition.
The \gls{csts} dataset includes 18,908 instances. Each instance contains a sentence pair, a condition, and a scalar for the similarity score on the 5-point Likert scale \cite{likert}. 
In this paper, we conduct the annotation and experiments on the \gls{csts} validation set that consists of 2,834 instances.
\citet{csts} evaluated the \gls{csts} dataset on different baselines such as SimCSE \cite{simcse} and GPT models \cite{gpt3,gpt4} by training or prompting with the sentence pairs and the conditions directly. However, our \gls{qa}-based method uses the generated answers as the model input. 

\paragraph{\gls{qa}-facilitated tasks}
Question answering tasks are useful for extracting and inferring information from the texts that is relevant to the question. 
Recent work utilized \gls{qa} to improve other NLP tasks. 
\citet{qa-sum1} and \citet{qa-sum2} applied \gls{qa} as an automatic evaluation metric for the summarization. \citet{qa-sum3} used \gls{qa} to improve the summarization directly.
Other works involved the application of \gls{qa} for data augmentation \cite{qa-aug} and question generation \cite{tu-etal-2022-semeval,cqg}.
In this paper, we apply \gls{qa} to generate condition-based answers for error identification and to improve models on the \gls{csts} task.

\begin{figure}[h!]
  \centering
  \includegraphics[width=1\linewidth]{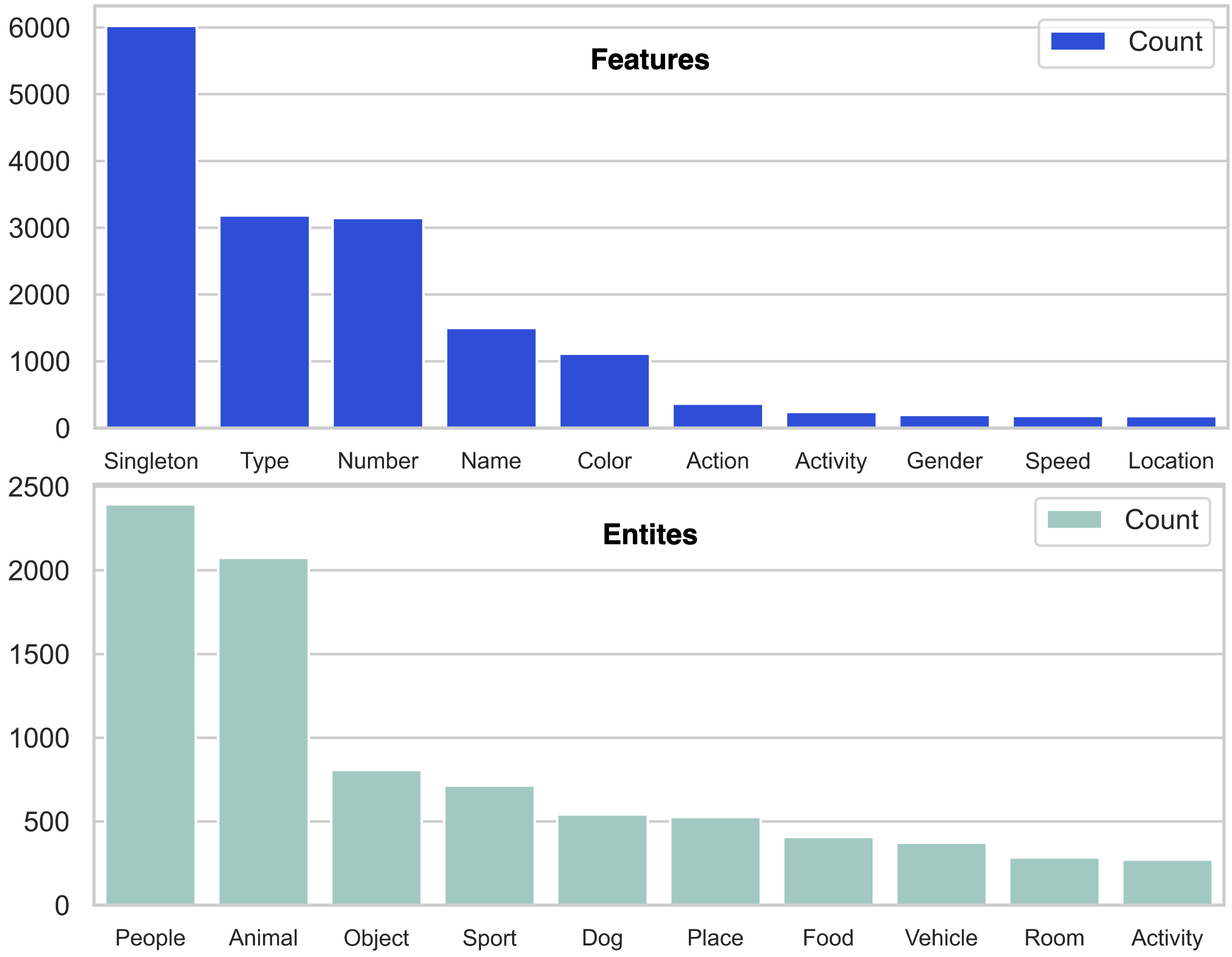}
  \caption{Distribution of top 10 frequent features and entities from the conditions in the \gls{csts} dataset. For the \texttt{singleton} with no explicit mention of the feature, we default the condition features from this group to \textit{type}.}
  \label{fig:condition}
\end{figure}

\begin{table*}
        \centering
        \resizebox{0.9\linewidth}{!}{
\begin{tabular}{l|l|l|l}

\textbf{Sentence Pair}                                         & \textbf{Condition} & \textbf{Label} & \textbf{Issue}\\ \hline
\begin{tabular}[c]{@{}l@{}}
\textcolor{brown}{Female tennis player}, standing on one foot after returned the ball.\\ 
\textcolor{brown}{A tennis player} is getting ready to hit the ball at a tennis match.\end{tabular} & number of people  & \textcolor{red}{3} / \textcolor{teal}{5} & Incommensurable Mapping \\ \hline
\begin{tabular}[c]{@{}l@{}}
An \textcolor{brown}{oak table} and chairs in a dining room, with a doorway to the kitchen.\\ 
Three couches positioned around a \textcolor{brown}{coffee table} in a living room.\end{tabular} & the table  & \textcolor{red}{1} / \textcolor{teal}{3} & Ambiguous Condition \\ \hline
\begin{tabular}[c]{@{}l@{}}
A \textcolor{brown}{toilet} in a stall with a sink attached and a console attached to the lid.\\ 
A \textcolor{brown}{bathroom} with tiled walls, a toilet, sink and a garbage can in it. \end{tabular} & room type  & \textcolor{red}{2} / \textcolor{teal}{4} & Inference Degree \\ \hline
\begin{tabular}[c]{@{}l@{}}
A \textcolor{brown}{cat} hissing as it attempts to fit itself into a bowl that it is to big to fit in.\\ 
A \textcolor{brown}{black cat} with a red tag sitting down with a bookshelf in the background.\end{tabular} & color of animal  & \textcolor{red}{2} / \textcolor{teal}{-1} & Invalid Condition \\ \hline

\end{tabular}}
        \caption{Examples with common issues that cause the judgment divergence between the \textcolor{red}{original} and \textcolor{teal}{reannotated} labels. \textcolor{brown}{Text} that is relevant to the conditions is highlighted.}
        \label{tab:errors}
    \end{table*}

\section{Reannotating \gls{csts}}
\label{sec:relabel}
We analyze the dataset and describe the annotation process for relabeling the \gls{csts} validation set.\footnote{The original labels of the \gls{csts} test set is not publicly available, so we use the validation split for further annotation and experiments in this paper.}

\subsection{Condition analysis}
We analyze the composition of the condition texts in the dataset.
We observe that the majority of the conditions are short phrases in the format of \textit{[feature] of [entity]} (e.g., \textit{the color of animals}) or simply a singleton \textit{[entity]} (e.g., \textit{the hobby}). We plot the distributions of frequent features and entities from the condition texts in the full \gls{csts} dataset in Figure \ref{fig:condition}, we notice that the dataset is skewed by having a long tail distribution of the conditions. The top 10 frequent features and entities appear in 88\% and 45\% of the total conditions respectively.
Within the top 10 lists, \textit{Type} and \textit{Number} are the dominating features; \textit{People} and \textit{Animal} are the dominating entities.

\subsection{Annotation Analysis}
To understand how conditions affect the human judgment of sentence similarity, we conduct a pilot annotation study on 150 instances sampled based on the frequency of condition features (e.g., \textit{type}, \textit{number}, etc.) in the \gls{csts} training set. Annotators are asked to reannotate those instances following the public \gls{csts} annotation guideline. We measure the agreement between the original labels and reannotated labels and find a low agreement of 40\% accuracy (exact label match) and 50.4 Spearman's Correlation.

We characterize the common issues that cause the annotation divergence in Table \ref{tab:errors}.
The similarity of the sentences under \textit{number} conditions cannot be mapped to a 5-point scale due to the binarity of the value comparison (e.g., 1 $=$ 1, 1 $\neq$ 5). This issue also happens to other condition features such as \textit{gender} and \textit{age}.
The condition can also be ambiguous, especially when it is a singleton. In the second example, the similarity between the two mentions of the \textit{table} can be subjective,  based either on the \textit{type of table} or multiple features associated with the table such as \textit{shape}, \textit{size}, etc.
The original \gls{csts} task does not specify how much inference from the context is allowed to judge the similarity. This increases the label inconsistency between the annotators. In the third example, although the room type is not explicitly mentioned in the first sentence, we can still confidently infer it is \textit{bathroom} because of the mention of \textit{toilet} and \textit{sink} in the context.
In the last example, the condition can be invalid if the information regarding the condition cannot be extracted or inferred from the sentence.

\subsection{Condition-aware Annotation on \gls{csts}}
We reannotate the \gls{csts} validation set to fix common annotation errors and resolve the aforementioned issues that cause the low agreement score.
The annotation was done by 4 researchers and graduate students from the linguistics and computer science departments of a US-based university.
Each annotator is familiar with the \gls{csts} annotation guideline and is well-trained through the trial annotation on the pilot set with 150 instances.
To resolve the issues that are identified from the pilot study, we ask annotators to follow additional annotation rules that are detailed as follows.
\paragraph{Incommensurable mapping} For binary conditions, only labels 1, 5, or 3 are permitted, representing equal, unequal and possible equal (e.g., comparing \textit{3} and \textit{several} in the \textit{number} conditions).
\paragraph{Ambiguous condition} Given the intuition that \textit{type} is always the primary feature in comparing the similarity between two entities, if conditions are singletons or have no features, we default it to \textit{the type of [entity]}.
\paragraph{Inference degree} Annotators are only allowed to make direct inference to the implicitly mentioned information with high confidence. For example, \textit{snow hill} indicates the weather, \textit{tennis} indicates the instruments being used, etc.
\paragraph{Invalid condition} We annotate invalid instances with the label -1 and exclude those instances from the reannotated dataset.

\begin{figure}[h!]
  \centering
  \includegraphics[width=1\linewidth]{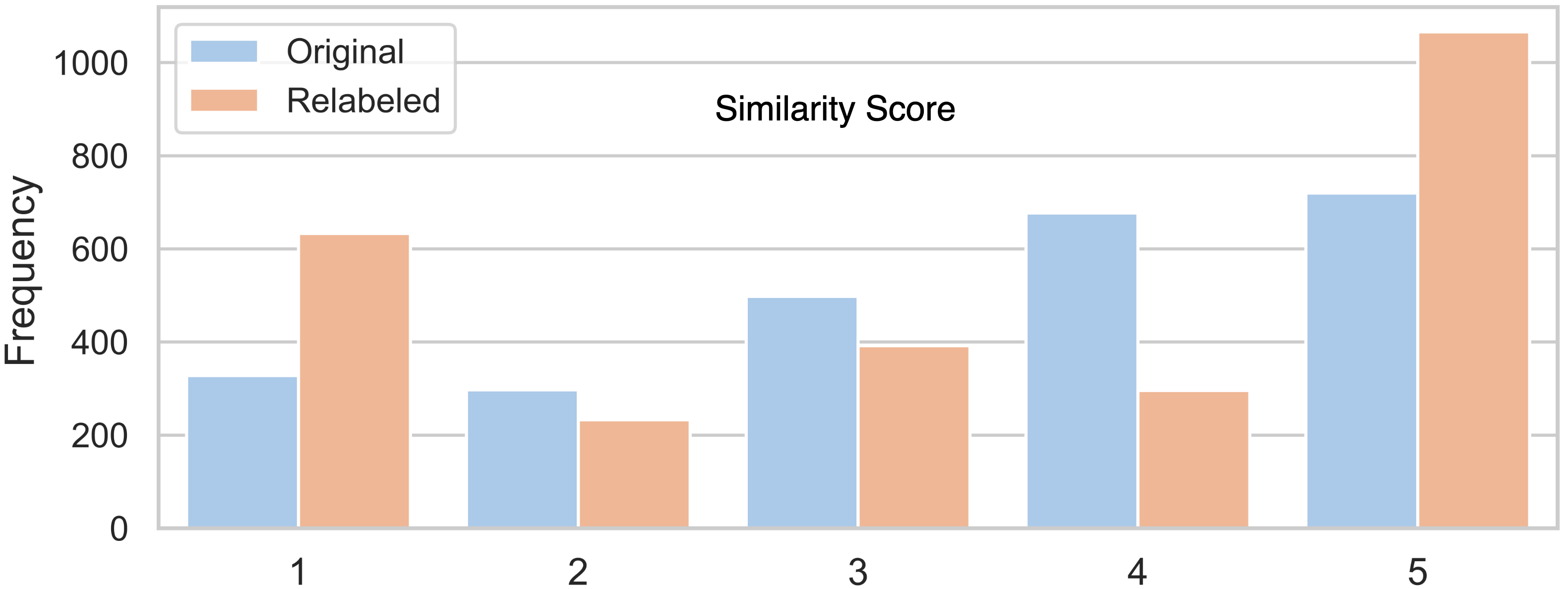}
  \caption{The similarity score distribution of the original and relabeled \gls{csts} validation set.}
  \label{fig:label_dist}
\end{figure}

After removing 214 instances with invalid conditions, we create a relabeled \gls{csts} validation set that consists of 2,620 samples.
Figure \ref{fig:label_dist} shows the label distribution of the original and relabeled \gls{csts} validation set.
Compared to the original set, the new annotation contains more extreme labels such as 1 and 5, while labels in between such as 2, 3, 4 are less frequent. This is due to the high frequency of the instances with binary condition features in the original dataset. 

\section{\gls{qa} for \gls{csts}}
With the consideration on scaling the relabeling task to the full \gls{csts} dataset, we explore effective approaches to identifying potential mislabeled instances automatically.
We apply \gls{qa} as a pre-task for identifying information from the sentences that is relevant to the condition, and leverage \gls{llms} to generate answers from condition-transformed questions.
We show that the generated answers correlate better to the reannotated labels, and can be used as a reliable intermediate resource for identifying potential annotation errors from the original \gls{csts} dataset. 

\begin{figure*}[h!]
  \centering
  \includegraphics[width=1\linewidth]{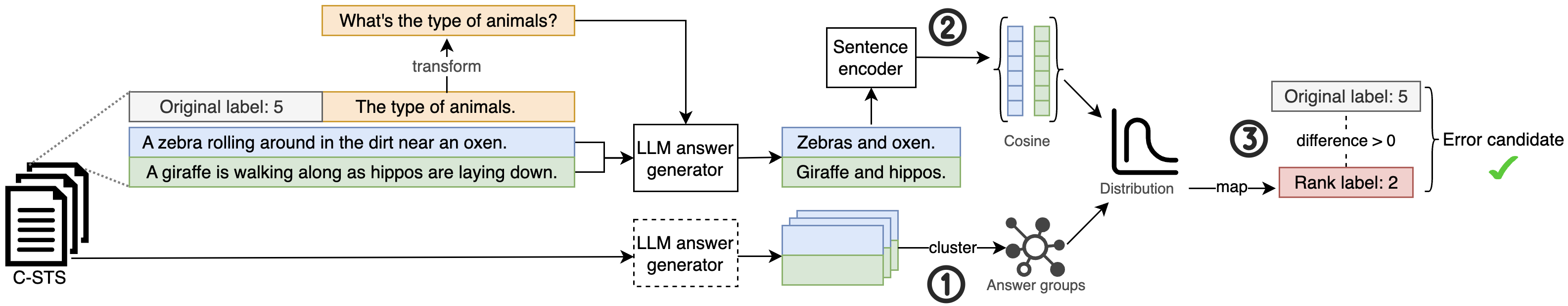}
  \caption{Answer generation and error identification pipeline on the \gls{csts} validation set.}
  \label{fig:qa_pipeline}
\end{figure*}

\subsection{Answer Generation}
\label{sec: answer_gen}
\paragraph{GPT prompting}
For each instance, we start by transforming its condition into a question with the format \textit{What is [condition]?}.\footnote{We convert the condition text to lowercase and remove the period at the end of the text.}
In order to generate high-quality answers, we conduct the \gls{qa} task with \gls{llms} under a prompting fashion.
Each prompt consists of a brief instruction, the original sentence and the condition-transformed question (Appendix \ref{app:prompt}). 
In the instruction, we ask the model to summarize each answer into a word or phrase to reduce potential noise and hallucinated content \cite{bouyamourn-2023-llms}. 
We experiment with GPT-3.5 \cite{gpt3} and GPT-4 \cite{gpt4} to generate answers. We use the OpenAI API with versions \texttt{gpt-3.5-turbo-1106} and \texttt{gpt-4-0125-preview}.

\paragraph{Answer quality analysis}
We evaluate the quality of the answers that are generated from the two models on 200 instances sampled from the \gls{csts} validation set.
We ask two annotators to measure the quality of the answer on a 5-point Likert scale from \textit{unrelated} to \textit{very accurate}. Answers from GPT-3.5 have a Spearman's Correlation of 4.54 and accuracy
of 75\% (labels with 5), While the Spearman and accuracy of GPT-4 answers are 4.07 and 69.5\% respectively.
We observe from the data that although GPT-4 is a more recent and proficient model, it tends to avoid generating direct answers with no clear instructions or contexts. Although this mechanism can help the model reduce hallucination, it misses what we consider as ``relevant answers'' in our task.

\enumsentence{
\small
{
\emph{A baby is swaddled and someone is putting his hands on it. [the age of child]} \\
GPT-3.5: \textbf{Infant}, GPT-4: \textbf{Not specified} }
\label{answer-exp1}}

\noindent
In example \ref{answer-exp1}, the answer from GPT-3.5 is \textit{infant}, which implies a rough range of the age, while GPT-4 refuses to generate an answer because the actual child age is not explicitly mentioned in the text.
Since our task involves measuring the text similarity of answers, we choose to use GPT-3.5 because it tends to generate more content-enriched answers than GPT-4. 

\paragraph{Answer correlation analysis}
Given the high quality of the GPT-generated answers, we evaluate the correlation between the answers and the reannotated labels.
We first run GPT-3.5 to generate the answers for all the instances from the \gls{csts} validation set.
Then we encode each answer into an embedding using the GPT embedding encoder with version \texttt{text-embedding-ada-002}, and compute the cosine similarity between the embeddings of the answers from the same instance.
We calculate the Spearman's Correlation between the answer similarities and reannotated labels (55.44), and between the original labels and reannotated labels (49.22).
The result shows that the cosine similarity of GPT-generated answers correlates more closely with the reannotated labels, and thus can better reflect the similarity of the sentence pairs on the given condition.

\subsection{Error Identification}

With the analysis that answers generated from GPT-3.5 are of high quality and correlate better with the reannotated labels, we propose a method to automatically identify potential annotation errors from the \gls{csts} dataset.
As illustrated in Figure \ref{fig:qa_pipeline}, the generated answers are the input to our error identification pipeline that consists of three steps: (1) clustering answers into groups with different topics; (2) encoding and ranking answer pairs in each cluster and mapping the similarity ranks to the label from 1 to 5; (3) identifying error candidates based on the difference between original and new rank labels.



\paragraph{Answer clustering}
Most of the generated answers can be grouped into different topics (e.g, answers related to \textit{numbers}, \textit{colors}, etc.), we apply K-means \cite{Arthur2007kmeansTA} to cluster the answer pairs. Each answer pair is concatenated and then encoded into a single embedding for the clustering.
The purpose of this step is to rank the similarity of the answers more accurately by clustering similar answer pairs together.

\paragraph{Answer similarity mapping}
We encode each answer from a pair into an embedding and compute the cosine similarity between the two embeddings. Within each answer cluster,  We rank all the answer pairs based on their embedding similarity and map the ranking of each answer pair to a new label (called \textit{rank label}) on a scale from 1 to 5 based on a predefined ranking distribution.

\paragraph{Error candidate selection} 
We identify the potential error candidates by comparing the original labels to the rank labels.
For each answer pair, if its rank label is different from the original label, we choose this instance as a candidate.


\begin{table}
    \centering
    \resizebox{0.9\linewidth}{!}{
    \begin{tabular}{lrrrrr}
        \toprule
        \multirow{2}{*}{\shortstack{ Distribution}} & \multicolumn{2}{c}{No Cluster} & \multicolumn{3}{c}{Clustered} \\
        \cmidrule(lr){2-3} 
        \cmidrule(lr){4-6} 
        & F1 & Spear. & F1 & Spear. & Best K \\
        \midrule 
         Even & 74.3 & 52.9 & 74.5 & 60.8 & 3\\
         Original & 75.4 & 55.4 & 76.6 & 59.6 & 10\\
         Reannotated & 79.4 & 49.1 & \textbf{82.4} & \textbf{65.7} & 10\\
         \bottomrule
    \end{tabular}}   
    \caption{Error identification results with or without k-means clustering under different distribution approaches on the test split of the relabeled \gls{csts} validation set. The results under clustered are from the cluster number (Best K) selected in the range of 1-20 that yields the highest F1 score.}
    \label{tab:error_id_res}
\end{table}

\subsection{Evaluation}
We evaluate our error identification pipeline on the \gls{csts} validation set. 
We use half of the dataset to generate answer clusters and ranking distributions, and the other half to evaluate three ranking distributions with different number of clusters. For the even distribution, we map the sorted answer pair rankings evenly to rank labels from 1 to 5 (e.g, 20\% of lowest ranked instances has rank label 1). For the original or Reannotated label distribution, we map the rankings based on the distribution of original/reannotated labels. We use precision, recall, and F1 as the evaluation metrics.

\begin{table*}
\centering
\resizebox{1\linewidth}{!}{
\begin{tabular}{l|l|l|l}
\textbf{Sentence Pair} & \textbf{Condition}  &\textbf{Response} & \textbf{Issue}\\ \hline
\begin{tabular}[c]{@{}l@{}} 
A couple of small pieces of cake sitting on top of a white plate. \\ 
A paper plate with a sandwich and a slice of pizza, both partly eaten. \end{tabular} & the amount of plates  & 
\begin{tabular}[c]{@{}l@{}}Two small pieces of cake
\\ One plate \end{tabular}
& GPT hallucination \\ \hline
\begin{tabular}[c]{@{}l@{}}
A group of people on a hill looking a city and two people are flying a kite.\\ 
Man playing with kite far up in the sky, showing only the deep blue sky. \end{tabular} & the kite  & 
\begin{tabular}[c]{@{}l@{}}
Recreational activity \\ 
High-flying entertainment
\end{tabular} & Vague condition \\ \hline
\begin{tabular}[c]{@{}l@{}}
A plate full of cut salad with a fork and a glass full of cold drink with ice. \\ 
A dish which contains cauliflower and meat is on top of a wooden tray. \end{tabular} & presence of meat  & 
\begin{tabular}[c]{@{}l@{}}
No meat \\ 
Meat included
\end{tabular}
 & Semantic mismatch \\ \hline
\end{tabular}}
\caption{Examples with common issues that cause the unidentified annotation errors. }
\label{tab:unidentified errors}
\end{table*}

\subsubsection{Results}
We show the evaluation results in Table \ref{tab:error_id_res}. 
Our pipeline is effective in identifying potential annotation errors with the baseline F1 of 74.3, significantly higher than the 55\% error rate in the \gls{csts} validation set.
The distribution from the reannotated label performs the best among all three distributions, suggesting the label consistency in the relabeled data.
Comparing to no cluster, ranking within the answer clusters improves the most on the reannotated label distribution, boosting F1 from 79.4 to 82.4. 
As a positive byproduct, clustering can also improve the Spearman's Correlation between the rank labels and reannotated labels, indicating that answers clustered into optimal number of groups can help produce more accurate and correlated rank mapping of the similarity.
Overall, we show the effectiveness of our error identification pipeline. It can also be easily extended and applied to identify errors from the rest of the \gls{csts} dataset with the help from the generated answers.

\subsubsection{Analysis}

We briefly characterize the cases where annotation errors are not identified by the pipeline. The complete examples are shown in Table \ref{tab:unidentified errors}.


\paragraph{Incorrect answer}
The incorrectness of the generated answers can be caused by: (1) hallucination from the GPT model; (2) vague or invalid condition.
For example, in the sentence \textit{A couple of cake ... a white plate.}, the GPT wrongly answers \textit{two pieces of cake} to the condition \textit{the amount of plates}.
In another example, the condition \textit{the kite} is not specific enough for both annotators and the GPT model, so the generated answer is highly depended on subjectivity.
These mistakes lead to an inaccurate answer similarity and thus cause a misaligned rank label.

\paragraph{Semantic mismatch}
Error identification relies on the ranking of the answer pair similarity. However, answers even with the opposite meaning can have a high similarity in the embedding space.
In an example, the answers are \textit{No meat} and \textit{Meat included}, which are opposite to each other.
However, the embeddings of the two answers have a high similarity due to the overlapped token \textit{meat}.

\section{Experiments}
We propose a new method that can improve the \gls{csts} task by utilizing the LLM-generated answers.
Instead of using the sentence pairs and the conditions directly as the model input, our method decomposes \gls{csts} into two subtasks: generating answers that encode the essential semantic information about the condition, and learning the similarity score between the answer pair.
We evaluate our method and compare with baseline models \cite{csts} under both fine-tuning and prompting settings. 
We randomly select 70\% instances from the reannotated \gls{csts} validation set for training and the remaining 30\% for testing.

\begin{table}
        \centering
        \resizebox{0.9\linewidth}{!}{
        \begin{tabular}{llrr}
              \toprule
              Model & Method & Spearman & Pearson\\
              \midrule
            
            \multirow{4}{*}{\makecell{Bi- \\ encoder}} & 
             SimCSE\textsubscript{\textsc{base}} & 49.6& 48.5 \\
             & SimCSE\textsubscript{\textsc{large}} & 71.7 & 70.8 \\
             & QA-SimCSE\textsubscript{\textsc{base}} & 73.9 & 73.4 \\
             & QA-SimCSE\textsubscript{\textsc{large}} & \textbf{75.9} & \textbf{75.4} \\
             \midrule
             \multirow{4}{*}{\makecell{Tri- \\ encoder}} &              SimCSE\textsubscript{\textsc{base}} & 0.8 & 1.7 \\
             & SimCSE\textsubscript{\textsc{large}} & 12.8 & 13.2 \\
             & QA-SimCSE\textsubscript{\textsc{base}} & \textbf{73.9} & \textbf{73.4} \\
             & QA-SimCSE\textsubscript{\textsc{large}} & {73.4} & {73.1} \\
             \midrule
             \multirow{4}{*}{\makecell{Cross- \\ encoder}} &              SimCSE\textsubscript{\textsc{base}} & 37.2 & 38.7 \\
             & SimCSE\textsubscript{\textsc{large}} & 43.0 & 43.3 \\
             & QA-SimCSE\textsubscript{\textsc{base}} & 71.4 & 71.1 \\
             & QA-SimCSE\textsubscript{\textsc{large}} & \textbf{72.9} & \textbf{72.3} \\
             \midrule
            \multirow{2}{*}{\makecell{GPT-3.5}} &         
             Base Prompt & 9.1 & 13.5 \\
             & QA Prompt & \textbf{66.1} & \textbf{64.8} \\
             \midrule
            \multirow{2}{*}{\makecell{GPT-4}} &         
             Base Prompt & 64.2 & \textbf{63.3} \\
             & QA Prompt & \textbf{64.4} & {60.2} \\
             \bottomrule
        \end{tabular}} 
        \caption{Evaluation results on the test split of the \gls{csts} relabeled set. We compare our methods (QA-based) with baselines under different model settings.}
        \label{tab:res}
\end{table}

\subsection{Model Setup}
\paragraph{Fine-tuning models} We evaluate our method on three encoding configurations, cross-encoder, bi-encoder \cite{sbert} and tri-encoder \cite{csts}. 
Unlike the baselines that encode sentences directly, our method chooses to encode the \emph{answers} on all three encoding configurations.
Cross-encoder encodes the concatenation of the answer pair and condition all together, while bi-encoder concatenates the condition to each answer and encodes them separately.
Tri-encoder has three encoders that encode sentences and the condition all separately. 
We use supervised SimCSE \cite{simcse}, one of the best-performing embedding models as the base sentence encoder for all three encoding configurations. 
We fine-tune each model on the training set and evaluate on the testing set. We use Spearman's Correlation as the primary evaluation metric.

\paragraph{Prompting models}
We compare our method with LLM baselines under a zero-shot prompt learning setting. We evaluate the results on GPT-3.5 and GPT-4. Similar to the experiments on fine-tuned models, instead of asking about the sentences directly, we formulate the prompt by instructing the model to score the similarity between the answers regarding the condition (Appendix \ref{app:prompt}).
\subsection{Results}

We show the model results in Table \ref{tab:res}.
For all models, our method (QA-based) improves the baselines by a large margin (73.1 improvement on tri-encoder), indicating the usefulness of model learning with answers. 
Fine-tuned models generally perform better than LLMs models, suggesting that the \gls{csts} task is sensitive to the in-domain training.

Compared with the large version of the models, our method makes more improvement to the base models. For example in the bi-encoder setting, the improvement for large and base model baselines is 4.2 and 24.3 respectively. This indicates the effectiveness of the method, especially on models with smaller sizes and fewer parameters. This may be due to the answers being already encoded with relevant semantic information, thus reducing the reasoning complexity for the small models.
This similar pattern also applies to LLM baselines, where GPT-4 performs much better than GPT-3.5 on the base prompt. However on the QA prompt, GPT-3.5 achieves slightly better result than GPT-4. \gls{qa} transforms \gls{csts} into an easier sentence similarity task that enables the use of more cost or resource efficient models without harming the performance much.

\begin{figure}[h!]
  \centering
  \includegraphics[width=1\linewidth]{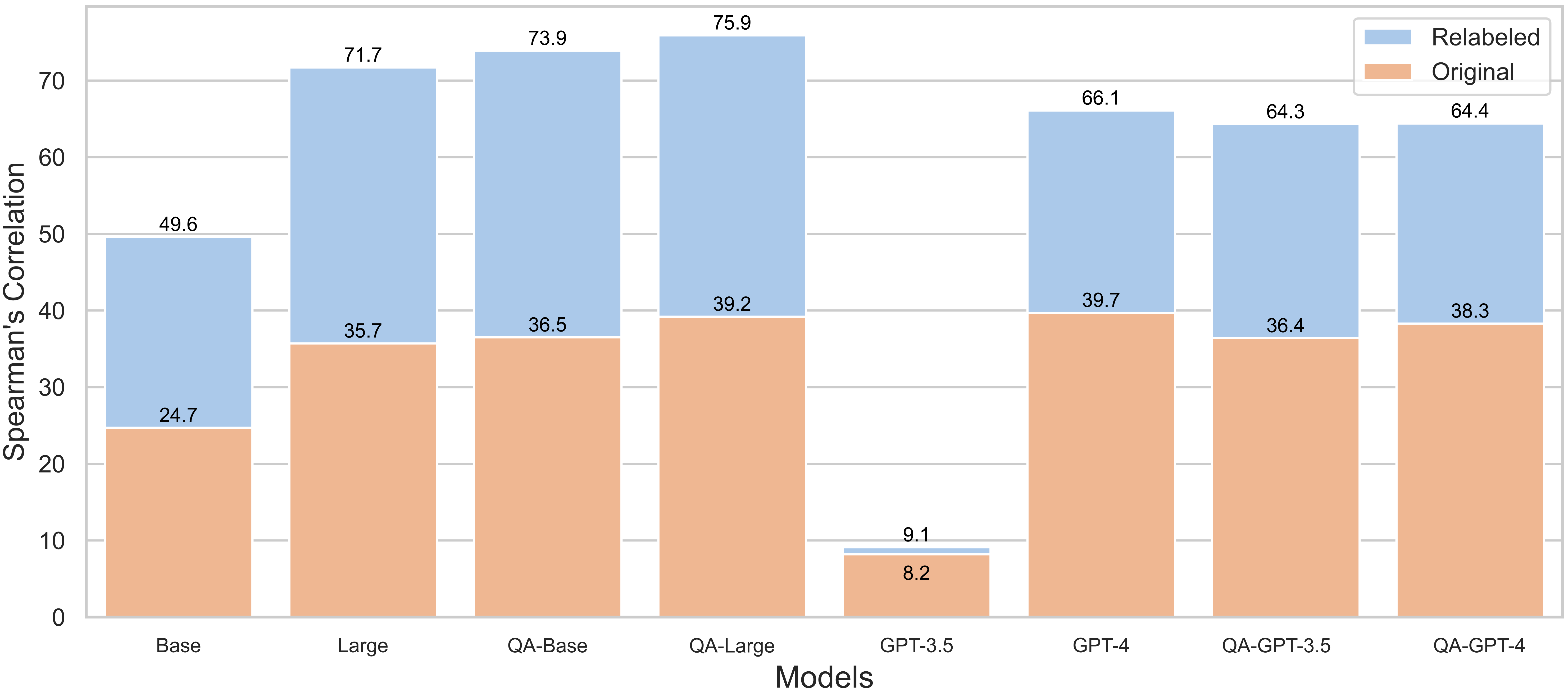}
  \caption{Model (SimCSE with bi-encoder and GPT) evaluation results on original and relabeled \gls{csts} validation set.}
  \label{fig:compare_relabel}
\end{figure}

\subsection{Analysis}
\paragraph{Curse from the mislabeled data}
We evaluate the effect of different label sets on the model performance. In Figure \ref{fig:compare_relabel}, we show the model results on the \gls{csts} validation set with original or reannotated labels. 
We notice that for all models, performance on the reannotated labels is increased by at least 40\% over the original label (except GPT-3.5).
This suggests that the original \gls{csts} dataset may not be able to truly reflect the capability of the existing models. The noise and vagueness from the original labels pose ``challenges'' to models to learn linguistic patterns.
Among the LLM baselines, we notice that \gls{csts} is particularly difficult to GPT-3.5 (9.1 Spearman). However, it can be improved significantly with our QA prompt.

\paragraph{Semantic information encoded with QA}
We evaluate how much semantic information can be encoded in the generated answers. As shown in Table \ref{tab:0shot-infer}, the condition is critical to the finetuned baselines (-0.4 to 49.6 on the SimCSE$_\textsc{base}$). However, training with no condition has minimal effect on the performance with the QA methods (2.4 difference on SimCSE$_\textsc{large}$).
Under the inference setting, the baselines perform poorly due to the additional reasoning complexity from the conditions. However, our method still shows strong performance even without any fine-tuning.
This echoes our previous finding that the QA subtask in \gls{csts} is able to improve the efficiency of model training .

\begin{table}
\centering
\resizebox{0.95\linewidth}{!}{
\begin{tabular}{lrrr}
              \toprule
              Method & Fine-tuning & No condition & Inference\\
              \midrule
            
             SimCSE\textsubscript{\textsc{base}} & 49.6 & -0.4 & 1.2 \\
              SimCSE\textsubscript{\textsc{large}} & 71.7 & 4.4 & 1.0 \\
              QA-SimCSE\textsubscript{\textsc{base}} & 73.9 & 73.3 & 49.4 \\
              QA-SimCSE\textsubscript{\textsc{large}} & \textbf{75.9} & \textbf{73.5} &\textbf{54.7} \\
             \bottomrule
\end{tabular}}  
\caption{Evaluation results (Spearman) from bi-encoder models under different training settings. \textit{Fine-tuning}: using models fine-tuned on the training data; \textit{No condition}: encoding answers only for fine-tuning; \textit{Inference}: inferencing results on the untuned models.}
\label{tab:0shot-infer}
\end{table}

\paragraph{Non-GPT answer generation performance}
We evaluate different answer generation models including multiple versions of Flan-T5 \cite{flant5}  and GPT-3.5.
We fine-tune the QA-SimCSE$_\textsc{base}$ on bi-encoder setting with answers generated from different models, and compare the results in Table \ref{tab:answer_model}.
Except Flan-T5$_\textsc{small}$, all the other models perform better than the baseline which is fine-tuned on sentences only.
The results suggest that our method is robust and does not necessarily rely on proprietary models such as GPT-3.5. Flan-T5$_\textsc{xl}$ is able to achieve comparable results to GPT-3.5 with only 1.5\% of the parameter size.
Even the base version of Flan-T5 with 248M parameters performs better than the baseline.

\begin{table}
        \centering
        \resizebox{0.9\linewidth}{!}{
        \begin{tabular}{lrrr}
              \toprule
              Generation Model & Size (Param.) & Spearman \\
              \midrule
            
             Flan-T5\textsubscript{\textsc{small}} & 77M & $\downarrow$43.7 \\   
             \midrule
             Baseline\textsuperscript{\ddag} & N/A & 49.6  \\
             \midrule
             Flan-T5\textsubscript{\textsc{base}} & 248M & $\uparrow$53.9  \\
             Flan-T5\textsubscript{\textsc{large}} & 783M & $\uparrow$55.8  \\
             Flan-T5\textsubscript{\textsc{xl}} & 2.75B & $\uparrow$62.3  \\
             GPT-3.5 & 175B & $\uparrow$66.1  \\ 
             \bottomrule
        \end{tabular}}   
        \caption{Evaluation results from QA-SimCSE$_\textsc{base}$ bi-encoder with difference answer generation models. \ddag: Fine-tuning on sentences only.}
        \label{tab:answer_model}
\end{table}

\section{Discussion: Improving Conditionality}
Although the idea of \gls{csts} is highly appreciated, we notice that the current \gls{csts} dataset has certain issues ranging from the errors in the annotation to the lack of rigor in the condition definition. 
In this section, we discuss a new annotation specification of conditionality based on a word's lexical attribute value features, or Typed Feature Structure (TFS)
\cite{carpenter1992logic,copestake2000definitions}, and exemplify the annotation of the new conditionality on several sentence pairs from the \gls{csts} dataset.

\subsection{\gls{tfs} as the Condition}
\begin{figure}[h!]
  \centering
  \includegraphics[width=0.9\linewidth]{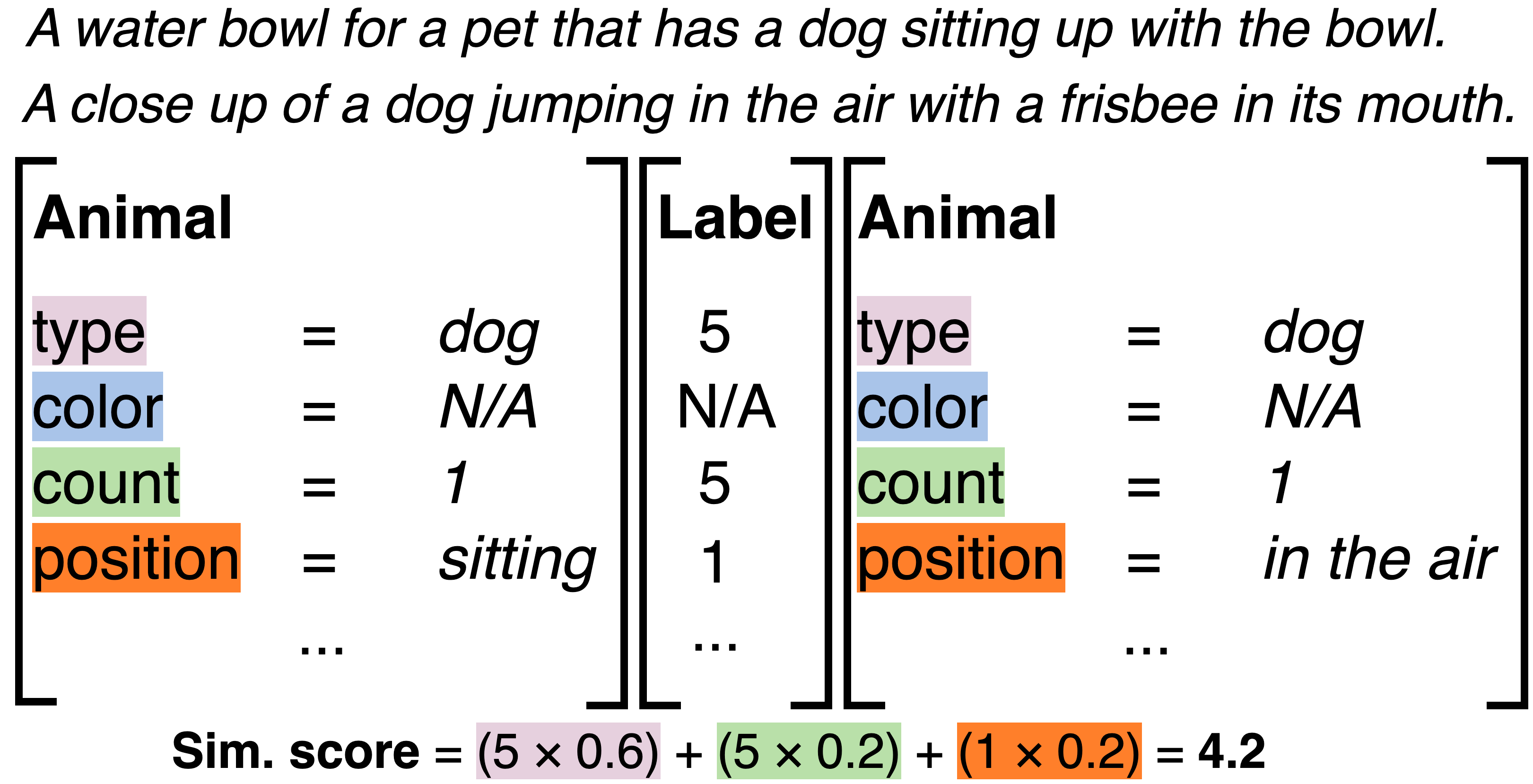}
  \caption{Illustration of the feature structures for the condition \textit{Animal}. Similarity score is calculated from the weighted sum of the similarity label for each feature.}
  \label{fig:tfs_ex}
\end{figure}

A \gls{tfs} is a data structure that can be used to represent systematic linguistic information about both words and phrases in language. For lexical items, the feature structure is defined as a set of attributes and their values for a word type. Each feature can have a distinct value for an object of that type \cite{pustejovsky2019lexicon}.
For example, in the term \textit{small table}, the value of the feature \textsc{size} for the object \textit{table} is \textit{small}.
In order to adopt the \gls{tfs} to construct conditions in the \gls{csts}, we use the word/entity type as the condition, and use the values of feature structure to annotate the similarity score regarding the condition. 
Consider Figure \ref{fig:tfs_ex} as an example. We set the entity type \textit{Animal} as the condition, and create a feature structure for each of the sentences. The final similarity score is calculated from the weighted sum of the individual similarity label annotated for each non-empty feature. The feature \textit{type} is the primary feature that contributes the most to the final similarity score.

It is worth noting that \gls{tfs}-based conditionality is highly extendable and customizable depending on the annotation needs. The condition can be selected from any node in a linguistic type hierarchy (e.g., \textit{animal} to \textit{mammal} to \textit{dog} in WordNet \cite{miller-1994-wordnet}). Correspondingly, the feature set for each condition can also be identified from existing lexical resources such as Schema.org\footnote{\url{https://schema.org/docs/about.html}} and ConceptNet \cite{Speer2016ConceptNet5A}.
Lastly, one can also decide how much weight from each feature needs to be assigned for calculating the final similarity score.
We leave the details on the \gls{tfs} design to future research.

\subsection{Qualitative Analysis}
We apply the \gls{tfs} to annotate example sentence pairs. 
\gls{tfs} conditionality enables a finer-grained mapping from entity semantic similarity to the final label, and mitigates the binarity from the evaluation on a single condition feature. 
\enumsentence{
\small
{
S1: \emph{On the plate there is croissant sandwich ...} \\
S2: \emph{... next to a hamburger on a green plate.} \\ 
Condition: \textbf{the food}, Score: \textbf{2.9}}
\label{tfs_ex1}}

\noindent
Consider example \ref{tfs_ex1}. 
Although the {type} of food is different (\textit{croissant sandwich} v.s. \textit{hamburger}), other features have high similarity: both food items are in the \textit{sandwich} category (i.e., category is the supertype of the type of the food) and held in a \textit{plate}; the amount of food are both \textit{one dish}.
While most of the entity features have similar or identical values, the primary feature balances it out and maps the final score to a relatively low similarity. 

\gls{tfs} also improves the clarity and objectivity of the conditions. Even when the entity type is the same, other feature values inferred from the sentence are able to differentiate the entities. 

\enumsentence{
\small
{
S1: \emph{A little girl is posing with a baseball bat ...} \\
S2: \emph{A kid is holding a baseball in a glove ...} \\ 
Condition: \textbf{the activity}, Score: \textbf{3.8}}
\label{tfs_ex2}}

\noindent
In example \ref{tfs_ex2}, both sentences indicate the type of the activity is \textit{baseball}, but participant and instrument are different: the first sentence mentions \textit{a girl with a bat}; while the second one omits the gender of the \textit{kid} and the instrument is \textit{glove}.


\section{Conclusion}
In this paper, we made a comprehensive analysis and improvement to the \gls{csts} task. With the reannotation effort on the original \gls{csts} data, we identified and resolved annotation errors and discrepancies that could hinder the evaluation of the task.
We showed that \gls{csts} can be naturally treated as a two-step reasoning task. We applied \gls{qa} to accomplish the first reasoning step by producing high-quality and correlated answers, and showed that the generated answers can be used effectively to automatically identify annotation errors and improve the \gls{csts} task under both supervised and prompting model settings.
Finally, we proposed to use the typed-feature structure in \gls{csts} to construct more semantically informed conditions. 
We believe that our work has led to a better execution of the \gls{csts} task. We hope that our analysis and improvement on the \gls{csts} can facilitate further developments by future researchers.

\section*{Limitation}
We reannotate the \gls{csts} validation set and show that our method can improve the model performance on the test split of the validation set.
We primarily use our reannotated labels as the \textit{gold} for the analysis and modeling. Although we have four annotators who are well trained for this work, there might still exist unintentional errors or subjective judgment. 

Due to the limitations of the resources, we could not scale the manual annotation to the full dataset by the time the paper is written. Finally we decide to use the validation set because the original labels for the test set are not public.
However, our dataset analysis (\textsection{\ref{sec:relabel}}) on the 150 instances from the train set does show that the errors and issues do exist in the whole \gls{csts} dataset. We believe that our error identification method can facilitate a more efficient reannotation work, and our modeling results and \gls{tfs}-based conditionality can be generalized to the rest of \gls{csts} data.

\section*{Acknowledgements}

This work was supported in part by NSF grant 2326985 to Dr. Pustejovsky at Brandeis University.
Thanks to Yifei Wang and Zhengyang Zhou for providing mathematical background knowledge on the error identification section of the paper.
Thank you to Jin Zhao for all the support. 
Thanks the reviewers for their comments and suggestions. The views expressed herein are ours alone.

\bibliography{custom}

\appendix
\section{Appendix}
\label{sec:appendix}

\subsection{Prompts}
\label{app:prompt}

\begin{figure}[h!]
  \centering
  \includegraphics[width=1\linewidth]{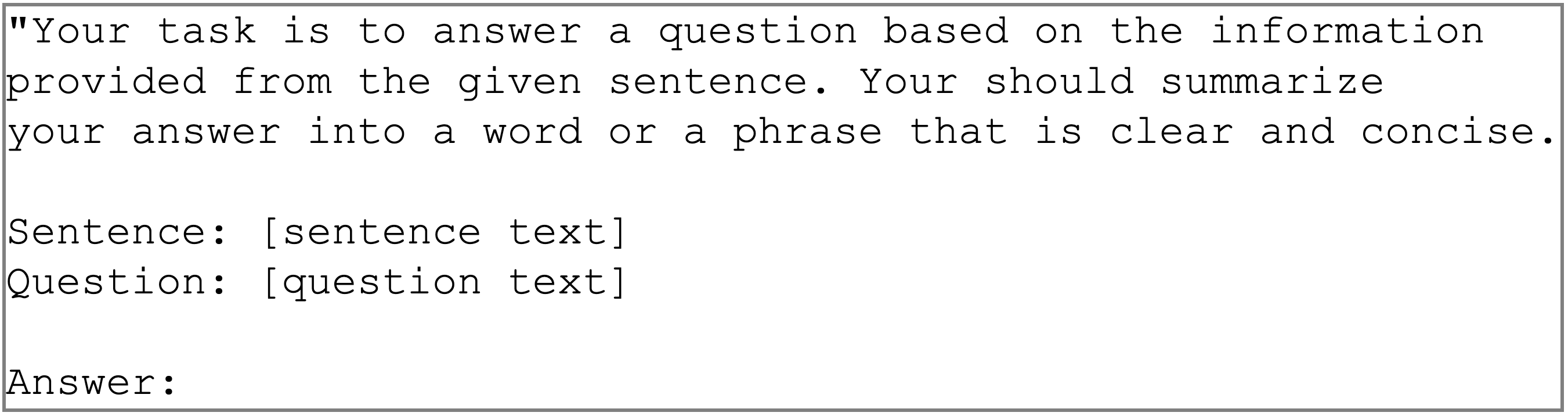}
  \caption{GPT prompt for the answer generation.}
  \label{fig:qa-prompt}
\end{figure}

\begin{figure}[h!]
  \centering
  \includegraphics[width=1\linewidth]{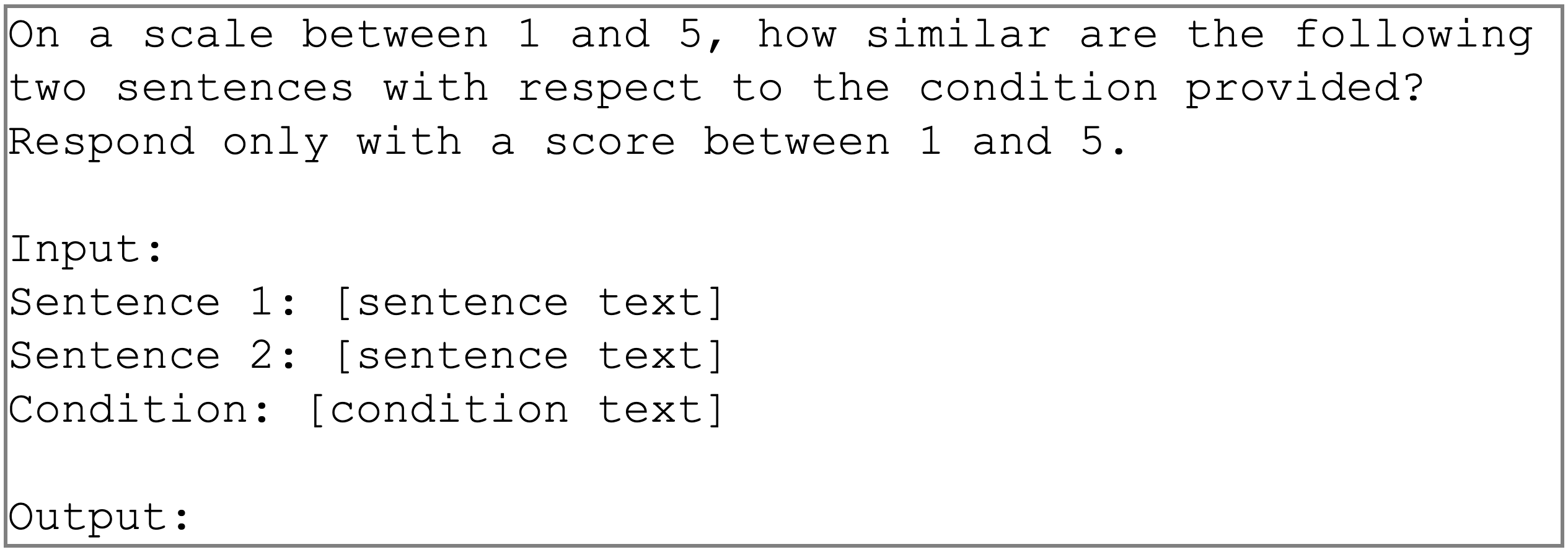}
  \caption{GPT prompt for the LLM baselines that is adopted from \cite{csts}.}
  \label{fig:baseline-prompt}
\end{figure}

\begin{figure}[h!]
  \centering
  \includegraphics[width=1\linewidth]{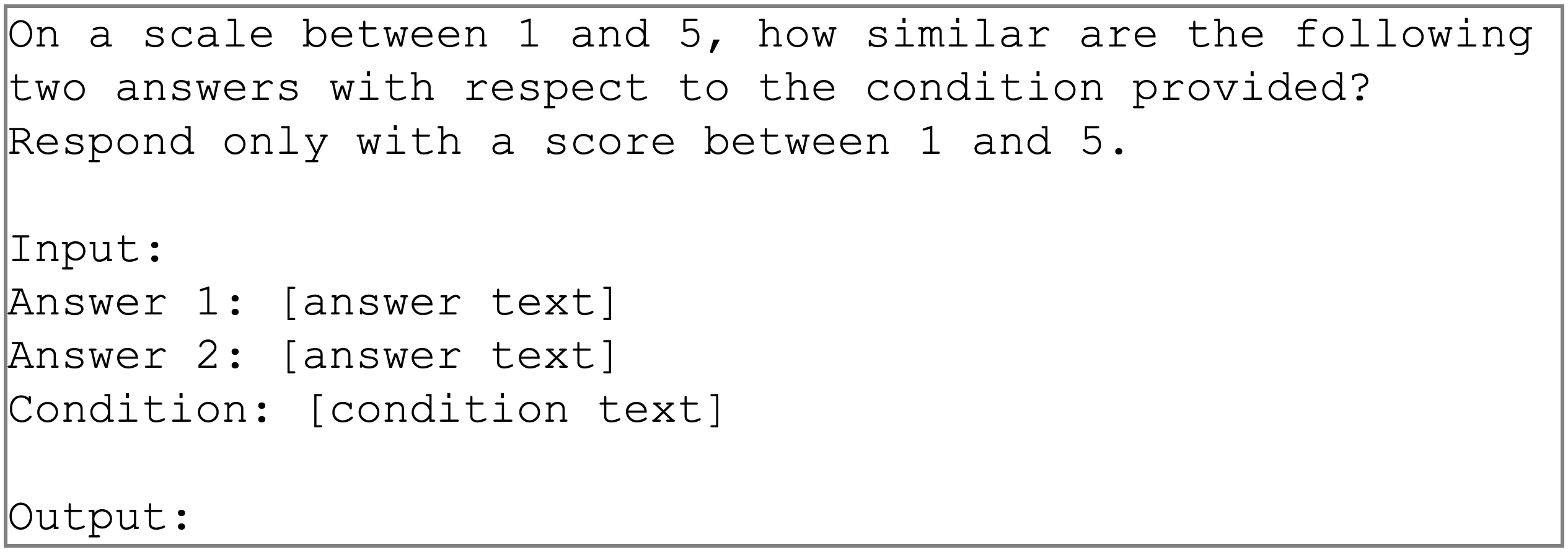}
  \caption{GPT prompt for our QA method on LLM.}
  \label{fig:qa-csts-prompt}
\end{figure}

We include various prompts that we used for the experiments in this section. Figure \ref{fig:qa-prompt} shows the prompt for the answer generation on GPT and Flan-T5 models.
Figure \ref{fig:baseline-prompt} and Figure \ref{fig:qa-csts-prompt} show the prompts for LLM baselines and our method. We instruct the LLMs to score based on answers instead of questions.

\subsection{Answer Clustering}
\label{app:answer-cluster}

\begin{figure}[h!]
  \centering
  \includegraphics[width=1\linewidth]{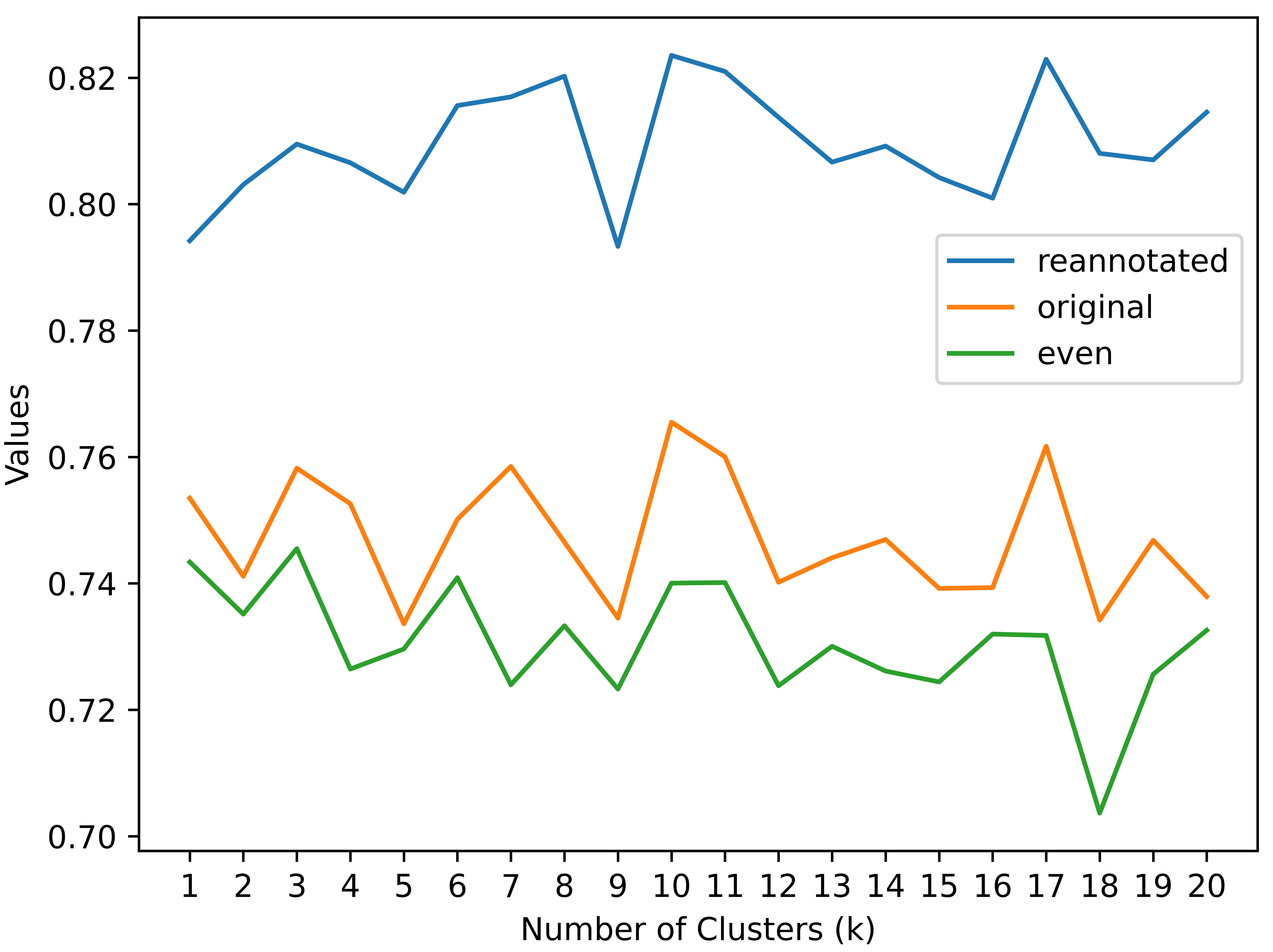}
  \caption{Error identification performance (F1 score) with different number of clusters on three distribution methods.}
  \label{fig:cluster-line-chart}
\end{figure}

We plot the change of the error identification results with different number of clusters in Figure \ref{fig:cluster-line-chart}. The reannotated distribution consistently performs better than the other two. The optimal cluster number is 10, 10, 3 for reannotation, original, and even distributions respectively.

\begin{table}
        \centering
        \resizebox{1\linewidth}{!}{
        \begin{tabular}{rrrrr}
              \toprule
               1 & 2 & 3 & 4 & 5 \\
               food & various topics & number & color & object \\
              \midrule
               6 & 7 & 8 & 9 & 10 \\
               gender, animal & location & activity & room related & activity \\
             \bottomrule
        \end{tabular}}   
        \caption{Topic from answer Clusters.}
        \label{tab:topics}
\end{table}

\begin{table}
        \centering
        \resizebox{\linewidth}{!}{
        \begin{tabular}{lrr}
              \toprule
              Model & Input & Output\\
              \midrule
            
             \texttt{gpt-3.5-turbo-0125} & \$0.0005 / 1K tokens & \$0.0015 / 1K tokens \\
             \texttt{gpt-4-0125-preview} & \$0.01 / 1K tokens & \$0.03 / 1K tokens \\
             \texttt{text-embedding-ada-002} & \$0.00010 / 1K tokens & N/A \\
             
             \bottomrule
        \end{tabular}}   
        \caption{Pricing of GPT models used in the paper.}
        \label{tab:price}
\end{table}


We show the answer topics identified from the clusters in Table \ref{tab:topics}. We use the cluster setting that produces the best result when the number of clusters is 10 and summarize the topics from each cluster. Most clusters capture the topics or patterns uniformly from the answers, such as \textit{color} and  \textit{food}. A few clusters have a mix of topics.


\subsection{Model Details}
\label{app:model}
We use OpenAI API to run GPT models. The pricing for the models used in the paper is shown in Table \ref{tab:price}.
We fine-tune SimCSE$_\textsc{base}$ and SimCSE$_\textsc{large}$ with different encoding configurations on a single Titan Xp GPU. We use the same hyperparameter setting with the baseline models in \cite{csts}. The training time is less than 10 minutes.
We use Flan-T5 with different sizes for inference only. We run Flan-T5$_\textsc{small}$ and Flan-T5$_\textsc{base}$ on CPU machines, and run Flan-T5$_\textsc{large}$ and Flan-T5$_\textsc{xl}$ on a Titan Xp GPU.





\end{document}